\documentclass[10pt,conference,a4paper]{IEEEtran}
\usepackage{booktabs}
\usepackage{ifpdf}
\usepackage{bm}
\usepackage{all-packages}
\usepackage{cite}
\usepackage{multirow}
\usepackage{cases}
\usepackage{amsmath}
\usepackage{array}
\usepackage{float}
\usepackage{fixltx2e}
\usepackage{stfloats}

\begin{document}

\title{A generalizable saliency map-based interpretation of model outcome}

\author{\IEEEauthorblockN{Shailja Thakur}
\IEEEauthorblockA{Electrical and Computer Engineering\\
University of Waterloo\\
s7thakur@uwaterloo.ca}
\and
\IEEEauthorblockN{Sebastian Fischmeister}
\IEEEauthorblockA{Electrical and Computer Engineering\\
University of Waterloo\\
sfischme@uwaterloo.ca}}

\maketitle

\begin{abstract}
One of the significant challenges of deep neural networks is that the complex nature of the network prevents human comprehension of the outcome of the network. Consequently, the applicability of complex machine learning models is limited in the safety-critical domains, which incurs risk to life and property. To fully exploit the capabilities of complex neural networks, we propose a non-intrusive interpretability technique that uses the input and output of the model to generate a saliency map. The method works by empirically optimizing a randomly initialized input mask by localizing and weighing individual pixels according to their sensitivity towards the target class. Our experiments show that the proposed model interpretability approach performs better than the existing saliency map-based approaches methods at localizing the relevant input pixels. 

Furthermore, to obtain a global perspective on the target-specific explanation, we propose a saliency map reconstruction approach to generate acceptable variations of the salient inputs from the space of input data distribution for which the model outcome remains unaltered. Experiments show that our interpretability method can reconstruct the salient part of the input with a classification accuracy of 89\%.

\end{abstract}



\acrodef{can}[CAN]{Controller Area Network}
\acrodef{ecu}[ECU]{Electronic Control Unit}
\acrodef{crc}[CRC]{Cyclic Redundancy Check}
\acrodef{dae}[DAE]{Denoising Autoencoder}
\acrodef{abs}[ABS]{Anti-Lock Brake System}
\acrodef{sdae}[SDAE]{Stacked Denoising Autoencoder}
\acrodef{dnn}[DNN]{Deep Neural Network}
\acrodef{gan}[GAN]{Generative Adverserial Network}
\acrodef{mse}[MSE]{Mean Squared Error}
\acrodef{svm}[SVM]{Support Vector Machine}

\newcommand{\probability}[1]{\mathrm{Pr}\left\{ #1 \right\}}
\newcommand{\fraction}[2]{\frac{\, #1 \,}{\, #2 \,}}
\newcommand{\Fraction}[2]{\displaystyle \fraction{#1}{#2}}
\newcommand{\norm}[1]{\left\lVert#1\right\rVert}
\def\inparenthesis#1{\left(#1\right)}
\newcommand{\vect}[1]{\boldsymbol{#1}}

\newtheorem{mydef}{Definition}

\def\implies{\Rightarrow}
\def\Implies{\Longrightarrow}
\def\iff{\Leftrightarrow}
\def\Iff{\Longleftrightarrow}
\def\assigned{~\leftarrow~}
\def\xor{\oplus}

\def\compromised{E_\mathrm{C}}
\def\target{E_\mathrm{T}}
\def\claimed{E_\mathrm{P}}
\def\predicted{\hat{E}}
\def\added{E_\mathrm{A}}

\newcommand{\bigOh}[1]{\mathrm{O}\left( #1 \right)}
\newcommand*\xbar[1]{%
   \hbox{%
     \vbox{%
       \hrule height 0.5pt 
       \kern0.5ex
       \hbox{%
         \kern-0.1em
         \ensuremath{#1}%
         \kern-0.1em
       }%
     }%
   }%
} 

\def\Section#1{Section~\ref{#1}}
\def\Figure#1{Figure~\ref{#1}}
\def\Eq#1{Equation~\eqref{#1}}
\def\Alg#1{Algorithm~\ref{#1}}
\def\Table#1{Table~\ref{#1}}
\def\Appendix#1{Appendix~\ref{#1}}
\ifx\undefined\FINAL

\def\proofingblankpage{\newpage}

\def\proofing#1{\textbf{\LARGE $\medstar$} \footnote{$\medstar\medstar$ \textbf{CM:}~~ #1 $\medstar\medstar$}}

\definecolor{Alarm}{rgb}{0.9,0,0}
\definecolor{Warning}{rgb}{0.8,0.3,0}
\definecolor{Good}{rgb}{0,0.75,0.25}

\def\brokensentence#1{{\color{Alarm} \proofing{Problem with sentence: #1}}}
\def\longsentence#1{{\color{Warning} \proofing{This sentence is too long; please rephrase}}}
\def\informal#1{{\color{Warning} \proofing{This sentence is too informal/coloquial: #1}}}
\def\praise#1{{\color{Good} \proofing{#1}}}
\def\unsubstantiated#1{{\color{Alarm} \proofing{Unsubstantiated claim: #1}}}
\def\incorrect#1{{\color{Alarm} \proofing{Factually incorrect: #1}}}
\def\careful#1{{\color{Warning} \proofing{Careful: #1}}}
\def\remove#1{{\color{Warning} \proofing{#1}}}
\def\change#1{{\color{Warning} \proofing{#1}}}
\def\warning#1{{\color{Warning} \proofing{#1}}}
\def\changed#1{{\color{Warning} \proofing{Changed it: #1}}}

\fi

\section{Introduction}
\label{sec:intro}

Recent advances in the field of deep neural networks have led to widespread applicability of artificially intelligent systems in the field of computer vision for the task of object detection~\cite{object-detection}, image classification~\cite{imagenet}, segmentation~\cite{segmentation}, image captioning~\cite{caption-generation}, visual question-answer~\cite{vqa}. Despite the significant advances in the speed and accuracy of neural networks, the complexity of the models makes the human-level understanding of the model's decision making a challenging problem. Notably, the highly non-linear interactions between the layers of the network make the outcome unintuitive and unpredictable. As a result of their inexplicable nature, their applicability remains limited in the domain of safety-critical systems (medicine, automotive, robotics, finance, nuclear) where a decision based on the outcome can lead to fatal consequences. 

Some of the research in the direction of explainable AI elucidate instances reflecting on the unpredictable nature of complex machine learning systems. For instance, in~\cite{lime}, the author shows how bias manifested in the machine learning algorithm through data leads it to misconstrued the characteristics of the snow for that of \textit{husky}. Another work by Stock et al.~\cite{basketball-bias-example} demonstrates the ImageNet~\cite{imagenet} bias introduced in the ResNet~\cite{resnet} model. As a consequence of the bias, the model prefers the image of a black person with a basketball for the class \text{basketball}, and Asians in red dress for the \textit{ping pong} class. Athalye et al.~\cite{turtle-rifle-example} show the sensitivity of the model can lead to misclassification. In the paper, the authors demonstrate that adding imperceptible perturbation to the input causes the model to misclassify the image of the turtle to the class of \textit{rifle}. 
 
All the above examples scenarios show the unpredictable nature of model prediction in the presence of uncertainty. Consequently, such advanced AI systems cannot be reliably used for decision making in critical systems that demand explanation and verification.

As a consequence of the black-box nature of complex AI systems, many possible solutions for understanding and interpreting the complex machine learning models have been developed in the last couple of years. One such technique is to visualize the activations of the individual layers of the network \cite{visualize-individual-layers}. However, for a particular image, this method is only able to tell apart what neurons are important for the classification of the input to a certain class. Saliency map-based methods exist \cite{visualize-saliency-map} that localize the input pixels which are sensitive towards the classification of the input to an output class. However, one of the limitations of these techniques is that they are intrusive; that is, they require access to the network parameters and gradients flowing through the network to localize the important input pixels. Thus, there is a need for a non-intrusive explanation technique for target-specific model outcomes.

\begin{figure}[!ht]%
\centering
\includegraphics[width=0.9\linewidth]{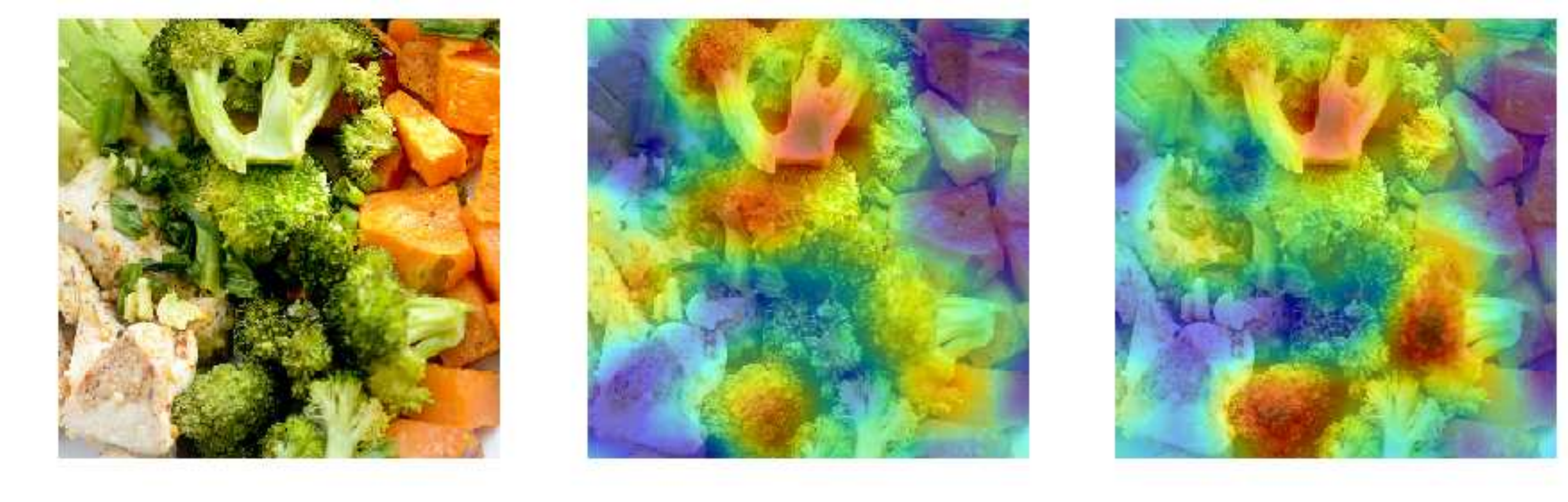}%
\caption{Saliency map generated for the image containing \textit{broccoli}. The two saliency map for \textit{broccoli} highlights non-overlapping regions of the image as important for \textit{broccoli} classification.}%
\label{fig:broccoli_preturbation_mask}%
\end{figure}

Furthermore, explanation using perturbation based techniques lacks consistency in the explanation. As shown in Figure~\ref{fig:broccoli_preturbation_mask}, multiple iterations of explanation of the same image result in contradictory salient region detection for the target class. Therefore, there is also a need for a global perspective on the explanation for a target class. 

\subsection{Contribution}
Our contribution to the paper is two-fold. First, we propose a non-intrusive interpretability technique by generating a saliency map based target-specific model outcome explanation. And second, we propose a method for generating \textit{alternate explanations} for the part of the input, which is salient for target-specific classification.

Inspired by the work of Petsuik et al.~\cite{rise}, we propose a non-intrusive explainability technique by generating a saliency map for a target class in less number of iteration ($N\sim1000$) than~\cite{rise} ($N\sim5000$). We use an \textit{empirical risk minimization} approach with a randomly initialized mask to locate the input pixels sensitive for the classification of the input to the target class. Therefore, if for the masked input, the confidence of the model in the most probable class is given by $p$, then the optimal set of pixels for the input is empirically located by randomly retaining $p$\% of the unmasked pixels (with value $> 0$) and $(1-p)$\% of the masked pixels (with value zero) followed by weighing the pixels using the class score. 
\begin{figure*}[!ht]%
\centering
\includegraphics[width=0.9\linewidth]{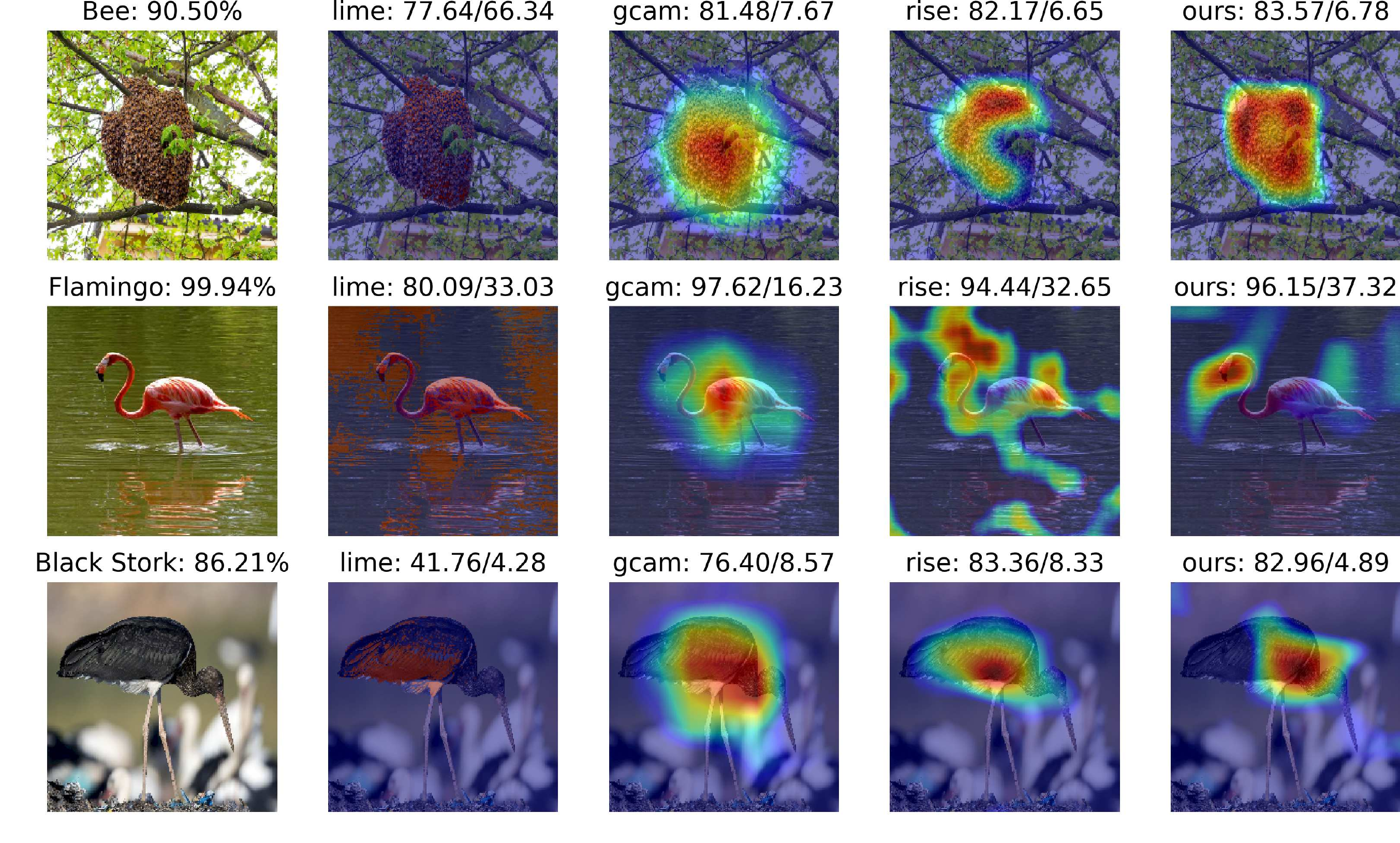}%
\caption{Saliency map generated for the target-specific image classification using our approach, RISE~\cite{rise}, GCAM~\cite{grad-cam}, and LIME~\cite{lime}. The first column shows the input image along with the top predicted class of the model outcome and the accuracy of classification. Second column onwards shows the saliency map overlapped with the input image and the AUC scores (\%) of insertion/deletion metrics~\cite{rise} where a higher value is considered good for insertion, and a lower value is considered good for deletion.}%
\label{fig:salince_map_comparison_intro}%
\end{figure*}

To generalize the explanation for the target class using the saliency map, we propose a technique to identify the variations of the pixels in the salient regions of the input for which the model prediction remains unaltered. The hypothesis for finding variations of the salient region comes from the analogy that the model is invariant to small perturbations in the input. Thereby, the approach helps identify variations (changes in colour intensities, object rotation, or inversion) in the salient region of the input space for which the model classification remains unaltered~\cite{invariance}. To generate alternative explanations for the salient regions of the input, we use an image completion technique proposed in~\cite{image-inpainting,context-encoder} that uses the features of the neighbouring pixels to reconstruct the pixels in the salient regions of the input. Using this approach, we are able to find an exhaustive and contextually similar set of transformations for the pixels in the semantic regions, which are classified to the same output class as the original input image.

\section{Related Work}
\label{sec:related-work}
With the increasing applicability of complex machine learning models, the need for an explainable and verifiable AI is increasing. In an attempt to justify models' outcomes, a variety of techniques have been proposed over the years. The majority of the explainability techniques fall into one of the three categories: (1) \textit{third-party explainer}: a separate model for explaining the outcomes of the base model  (2) justify the base model outcome using techniques such as input perturbation and network parameters.

Some of the explainability work that falls in the line of \textit{third-party explainer} include \cite{ multimodel-explanation, tcav}. In particular, \cite{third-party-visual-explain} uses the class discriminative properties of the objects in the images to provide a textual explanation for the images. \cite{multimodel-explanation} is another technique that trains two models for providing textual as well as visual justification for the visual question answering task and activity recognition task. However, these approaches are costly to implement because of the reliance on the availability of large human-annotated ground-truth explanations. 

Within the realm of justifiable models, a variety of explainability approaches have been developed. One of the earliest approaches~\cite{lime} attempts to provide a linear interpretation within the local neighbourhood of the data point. However, the approach is not effective at explaining non-linear models. Some of the approaches~\cite{visualize-saliency-map,visualize-individual-layers,input-synthesized,gbackprop} attempts to synthesize input images that result in a high activation score for particular neurons. Another approach by~\cite{cam} generates a target-specific saliency map by taking the global average pooling of the feature maps at the layer before the fully connected layer. GradCAM \cite{grad-cam} is a generalized version of CAM that, in addition to the feature map weights, feeds the class gradient to the fully connected layer to assign importance to each of the input pixels. However, \cite{cam,grad-cam} can only be applied to limited network architectures with global average pooling. Another work by Zhang et al.~\cite{top-down-neural-attention} proposes to use a backpropagation scheme to generate an attention map by propagating the signal downward through the network hierarchy using a winner-take-it-all strategy. A few techniques examine the relationship between input and output to learn a perturbation mask by backpropagating the error signal \cite{preturbation}.
  
Despite the ability of the techniques \cite{cam,grad-cam,preturbation,real-time-image-saliency} to justify the model's decision, the methods mentioned above have limitations. The methods~\cite{grad-cam} are constrained by the use of network parameters such as gradients flowing through the network and network layer weights. While techniques such as \cite{top-down-neural-attention,grad-cam} require a specific kind of network architecture, in some cases \cite{visualize-individual-layers}, the method requires access to intermediate layers of computation for visualizing the features at several layers. Furthermore, the techniques can explain only a particular input at a time, without taking into consideration the possible variants (rotation, inversion, deformations) of the image. A work by Kim et al.~\cite{tcav} proposes a technique to provide an explanation that is representative of user-defined concepts, but the manually generated concepts limit the technique. Our work is an extension of the work by Petsuik et al.~\cite{rise} to localize and generalize the salient pixels of the target class using a saliency map. We obtain a saliency map (using $N$ less than that of ~\cite{rise}) by empirically optimizing the pixels important for target-specific classification. And, we propose an approach to provide a global perspective on the explanation of the outcome using a reconstruction technique, which generates possible variations of the salient pixels of the input.

\section{Proposed Technique}
\label{sec:proposed-technique}

We propose a non-intrusive explainability technique by localizing and generalizing the parts of input important for target-specific classification. To localize the pixels sensitive for target-specific classification, we extend \cite{rise} to generate a saliency map. Second, to obtain a global perspective on the saliency map of the model outcome, we propose a salient region reconstruction approach that reconstructs the input image with alternate variations of the pixels from the salient region of the input, all of which classifies to the target class. 

\subsection{Saliency Map Generation}
Given an image $I$ of dimension $H \times W$ from the space of images $\mathcal{I}=\{I:\Lambda^{H \times W} \to \mathbb{R}^3\}$ that maps each pixel coordinate to three color values, a target class $c \in \mathcal{C}$, a classifier $f:\mathcal{I} \to \mathbb{R}^\mathcal{C}$, which maps inputs from the input space, $\mathcal{I}$ to a vector of real numbers signifying the strength of the classifier in the output classes, $\mathcal{C}$, a random initial mask, $M_0=\{\Lambda^{h \times w} \to [0,1], h<H, w<W \}$. $M_0$ is composed of a set of unmasked and masked pixels such that $\Lambda^{h \times w}=\{\Lambda_\text{on}^{h1 \times w1} \cup \Lambda_\text{off}^{h2 \times w2}\}$, and $n_1$ and $n_2$ are the number of pixels in the $\Lambda_\text{on}$ and $\Lambda_\text{off}$ sets. Thus, given a binary mask $M_0$, if $\lambda \in \Lambda_\text{on}$, then $M_0(\lambda)=1$ else $M_0(\lambda)=0$.  



We use the \textit{empirical risk minimization} approach to iteratively update $M_0$ by preturbing the maksed and unmasked pixels set $\Lambda_\text{on}$ and $\Lambda_\text{off}$ based on the prediction probability, $p=f(I \odot M, c)$ where $\odot$ is the element wise multiplication and $M$ is the upsampled version of the mask at the $i^{th}$ iteration. $M$ is upsampled using bilinear interpolation to the size of the input image $I$ as shown in the second column of Figure~\ref{fig:saliency-mask}. The bilinear interpolation is a common resizing technique in computer vision that helps avoid the inclusion of unwanted artifacts in the mask during empirical optimization by blurring out the edges, as shown in figure~\ref{fig:saliency-mask}, thus, eliminating misclassifications. 


\begin{algorithm}
\label{alg:saliency-map}
 \caption{Algorithm for generating saliency map}
 \begin{algorithmic}[1]
 \renewcommand{\algorithmicrequire}{\textbf{Input:}}
 \renewcommand{\algorithmicensure}{\textbf{Output:}}
 \REQUIRE Input image $I \in \mathcal{I}$, Target class $c \in \mathcal{C}$
 \ENSURE  saliency map $M$
 \\ \textit{Initialisation} : $M_0 \in \mathbb{R}^{h \times w}$  
  \FOR {$i = 1$ to $N$}
  \STATE $M =\text{Resize}(M_i)$ 
  \STATE Compute $p=f((I\odot M), c)$
  \STATE $\Lambda_1 \leftarrow $ Randomly select $n_1p$ pixels of $\Lambda_\text{on}$
  \STATE $\Lambda_2 \leftarrow $ Randomly select $n_2(1-p)$ pixels of $\Lambda_\text{off}$
  \STATE $\Lambda_\text{on} \leftarrow \Lambda_1 \cup \Lambda_2$
  \STATE $\Lambda_\text{off} \leftarrow \Lambda \setminus \Lambda_\text{on}$
  \STATE If $\lambda \in \Lambda_\text{on}$ then $M_i(\lambda) \gets M_i(\lambda)p$
  \STATE If $\lambda \in \Lambda_\text{off}$ then $M_i(\lambda) \gets 0$
  \STATE $V=\sum\limits_{x,y}\Vert M_{xy+1}-M_{xy}\Vert^2 +\sum\limits_{x,y}\Vert M_{x+1y}-M_{xy}\Vert^2$
  \STATE $M_i \leftarrow M_i + \eta \Delta V \Delta p$
  \ENDFOR
 \RETURN $M$ 
 \end{algorithmic} 
 \end{algorithm}


The algorithm~\ref{alg:saliency-map} shows the saliency map generation method. Based on the prediction probability of the target class $c$, the algorithm randomly retains $p\%$ of the unmasked pixels, $\Lambda_\text{on}$ and $(1-p)\%$ of the masked pixels, $\Lambda_\text{off}$. The subset of the pixels preserved from $\Lambda_\text{on}$ and $\Lambda_\text{off}$ forms the new set of unmasked pixels $\Lambda_\text{on}$, which are sensitive for the classification of input $I$ to the target class $c$. However, the mask update is likely to saturate if the change in the prediction probability $p$ of the target class is negligible. To overcome the issue of saturation, we add a regularizer, which penalizes the set of unmasked pixels according to two factors: the change in the prediction probability of the target $\Delta p$ and change in the total variation in the mask $\Delta V$ where $\Delta$ denotes the change in the value from the previous iteration. If the mask is invariant to small changes in the classification accuracy of the target, then the regularizer will penalize unmasked pixel values heavily by adding a large negative penalty. On the other hand, if the change in $\Delta p$ is significant, then the regularizer adds a low penalty to the updated mask. This way, $M$ captures the pixels which are sensitive towards the classification of $I$ to the target class $c$. 


\subsection{Variations of the salient region of the input}
\label{sec:alternate-explanations}
\begin{figure}[!ht]%
\centering
\includegraphics[width=1.0\linewidth]{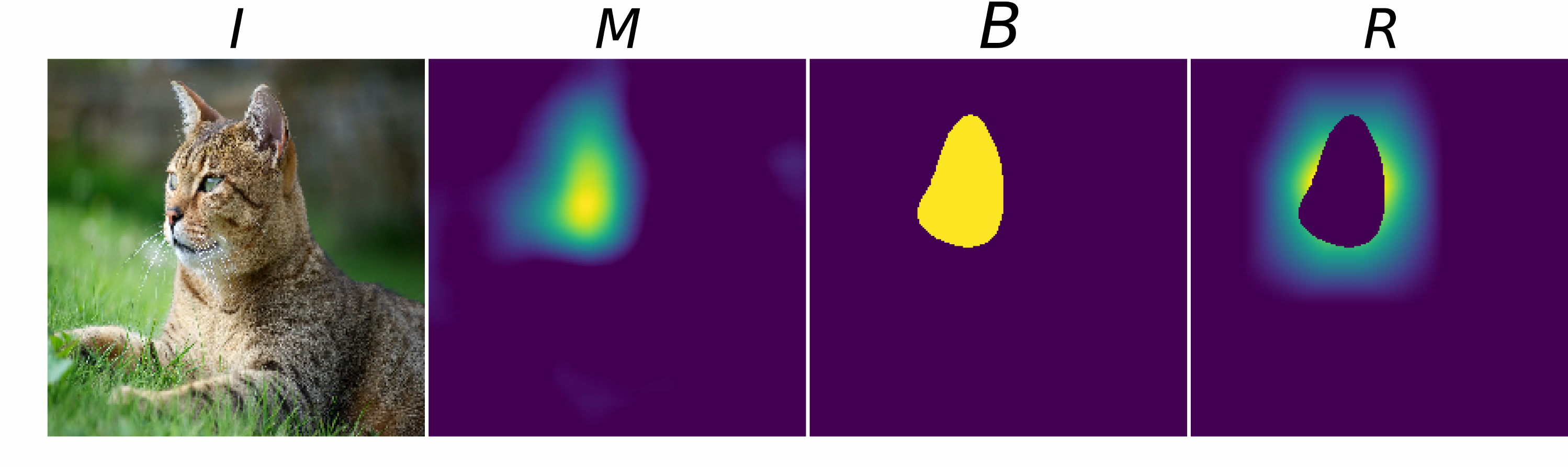}%
\caption{From left to right, \textit{I}: input image, \textit{M}: saliency map, \textit{B}: bounding box, \textit{R}: reconstruction mask.}%
\label{fig:saliency-mask}%
\end{figure}

We propose an approach to generate the acceptable variations for the salient region (the highlighted region in second column of Figure~\ref{fig:saliency-mask}) of the input image as \textit{alternate explanations} for the model outcome. 
To achieve this, we use an image completion technique proposed by Pathak et al~\cite{context-encoder}, which reconstructs a patch of the input image by iteratively backpropagating the error in a generator. 

Let $B$ be a binary bounding box for the saliency mask $M$ of the input image $I$, as shown in Figure~\ref{fig:saliency-mask} third from left, where pixels inside the box are set to one, and the pixels outside are set to zero. The reconstruction mask, $R$, is obtained by inverting $B$, followed by convolving $B$ with a kernel of size $(s,s)$ such that the weights assigned to pixels are inversely proportional to their distance from the bounding box. As pixels near the box are more important for the reconstruction of missing pixels from the box, the mask $R$ is such that it assigns more importance to the pixels in the vicinity of the box than to the pixels far away. Let $G$ be a generator that learns an encoding $d_z$ of the input image distribution $d_\mathcal{I}$ in the latent space ($z$). 

The objective is to reconstruct the corrupt image $((\bm{1}-R) \odot I)$ using $G$. As the corrupt image is not a sample from the input distribution $d_\mathcal{I}$; therefore, $G$ will be poor at recognizing the patterns of the missing part of the image. Therefore, we use the image recontrustion technique as described in~\cite{image-inpainting}, where the authors use the back-propagation technique with the generator to find an encoding ($z^\prime$) for the missing part of the image that is closest in encoding to the input $I$ while being confined to the learned manifold ($z$). The objective function for learning the encoding ($z^\prime$), comprises of context loss and discriminative loss.

\textbf{Context Loss} is used to reconstruct the missing part of the original image given the corrupt image by measuring a squared error between the corrupt image and the reconstructed image.

\begin{equation}
L_{cxt}(z) = (I \odot (\bm{1}-R)) - (G(z) \odot (\bm{1}-R))
\end{equation}

\textbf{Discriminative Loss} is used for measuring the authenticity of the generated images by feeding them to the discriminator $D$, which returns the confidence in $G(z)$ being real.

\begin{equation}
L_{dis}(z) = -D(G(z))
\end{equation}

The overall loss for learning the encoding for the missing salient region of the input is as follows,
\begin{equation}
L(z^\prime) = \operatorname*{argmin}_z \{L_{cxt}(\bm{z}|I, R) + L_{dis} (z)\}
\end{equation}

Using the learned encoding, G($z^\prime$) generates the image that is approximately close to the missing salient part of the image. The image reconstructed using the generated image is given by,

\begin{equation}
I_\text{rec} = (I \odot (\bm{1}-B)) + (B \odot G(z^\prime))
\label{eq:image-reconstruction}
\end{equation}

We repeat image reconstruction for $k$ randomly sampled noise vectors $z$ to generate $k$ reconstructed variants of the salient region of the input image $I$ where $z$ is a sample from a gaussian distribution with mean zero and variance one. The generated images will be such that their encoding will lie in the vicinity of the learned manifold in the latent space ($z$). However, if the encoding $z^\prime$ fails to capture the context of the salient region of the input using the evidence from the local neighbourhood of pixels, then some of the reconstructed images will not be contextually similar to the original input salient region. The selected set of $I_\text{rec}$ for which the model prediction remains unaltered are considered \textit{alternative explanations} for the salient regions of the input image. 

\section{Evaluation}
In this section, we give an overview of the evaluation metrics, models, and datasets used, and describe the results of the evaluation of the proposed approach. 

\subsection{Model and Data Description}

We evaluate the efficacy of the proposed saliency map based explainability approach using a range of publically available open image datasets such as ImageNet~\cite{imagenet} and MS-COCO (Microsoft-Common Object in Context)~\cite{mscoco}. ImageNet is a repository of 15\,M high-resolution images gathered from more than 20\,k categories. MS-COCO, on the other hand, is a significantly smaller database of images but with more number of instances per category. The dataset has 330\,k images with more than 80 object categories. The MS-COCO dataset is used for object detection, object segmentation, and image captioning. 

We use pre-trained models: VGG16~\cite{vgg}, Inception V3~\cite{inception}, and ResNet50~\cite{resnet} as base models for image classification. The pre-trained models are loaded with ImageNet weights and accepts inputs of size $224 \times 224$. The models differ in terms of their network structure and number of trainable parameters. VGG16 is a convolutional neural network trained on ImageNet dataset achieving top-5 accuracy of 92.0\% on ImageNet. ResNet50 is a 50 layer network achieving 93.29\% accuracy on the ImageNet. Inception V3 is another model that achieves an accuracy of 94.4\% on the ImageNet. Relying on pre-trained models~\cite{vgg, inception,resnet} for image classification helps avoid the common training pitfalls such as model over-fitting, skewed data distribution, right model selection, and insufficient resources such as GPUs for training. 

\subsection{Evaluation Metrics}
\subsubsection{Insertion and Deletion Metrics}
Motivated by~\cite{rise}, we use the metrics of insertion and deletion to evaluate our saliency map approach. In the deletion metric, the deletion of salient pixels from the input causes the model to drop the probability of the target class. And in the insertion metric, the insertion of pixels from the relevant region of the input causes the model to increase the probability of target class. We capture the sensitivity of the model to the removal and insertion of pixels from the relevant region of the input using an average AUC (Area Under the Curve) score. Thus, during deletion, as the relevant input pixels are deleted from the masked input, the AUC curve for the model will shrink to a thin area, thus, dropping the average AUC score, indicating the right explanation for the model decision. Similarly, during the insertion, as the pixels from the relevant region of the input are added to the masked input, the AUC curve expands to cover the large area under the probability curve, thus increasing the average AUC score.

 \subsubsection{Pointing Game}
The pointing game~\cite{top-down-neural-attention} metric is a method of evaluating the class discriminative nature of saliency-map based approaches. Given an annotated segmentation box for an instance of an object and the corresponding saliency map, the method measures the number of pixels on the saliency map of the input that lies on the annotated box of the object instance. To measure the overlap, we calculate the fraction of the area ($A_s$) of the saliency map that overlap with the annotated segmentation box ($A_t$) of the image using an IOU (Intersection Over Union) score (\%) = $\frac{\sum(A_s \cap A_t)}{\sum(A_s \cup A_t)}$. The numerator is the sum of pixels values in the union of $A_s$ and $A_t$, and the denominator is the sum of pixels values in the intersection of $A_t$ and $A_s$. A high IOU score (typically $\geqslant 50\%$) means that a large fraction of the salient region of the input overlaps with the annotated box, indicating a good explanation. 

\subsection{Evaluation using insertion and deletion metrics}
Given a pre-trained classifier, we evaluate the class discriminative capability of saliency-map based approaches~\cite{lime,grad-cam,rise,gbackprop} using the quantitative measures of insertion and deletion metric. For the base models: VGG16~\cite{vgg}, ResNet50~\cite{resnet}, and Inception V3~\cite{inception} and the datasets: ImageNet~\cite{imagenet} and MS-COCO~\cite{mscoco}, we report the average AUC score of insertion and deletion on a set of images. The test images are randomly selected from the test set of the datasets. For each technique, Table~\ref{tab:evaluation-inser-del} shows the mean AUC score of the insertion and deletion metrics for each technique across all the models and all the datasets. We also show the standard deviation of the AUC of the insertion and deletion metrics for our approach. From the table, it is evident that our approach outperforms other saliency-map based approaches in localizing the pixels sensitive for the classification of the input to the target class across both the datasets and all the base models. 

 \begin{table*}[ht]
    \centering
    \caption{Mean AUC Score(\%) using insertion (Ins) and deletion (Del) metrics}
    \begin{tabular}{lccccccccccc}
    \toprule
    \multirow{2}{*}{Model} & \multirow{2}{*}{Dataset} & \multicolumn{2}{c}{Ours} & \multicolumn{2}{c}{RISE~\cite{rise}} & \multicolumn{2}{c}{GCAM~\cite{grad-cam}} & \multicolumn{2}{c}{LIME~\cite{lime}} & \multicolumn{2}{c}{Guided Backprop~\cite{gbackprop}}\\
    
                           & & Ins & Del & Ins & Del & Ins & Del & Ins & Del & Ins & Del  \\
    \midrule
    \renewcommand{\arraystretch}{1.2}

    \multirow{2}{*}{ResNet50~\cite{resnet}} & MS-COCO~\cite{mscoco} & \textbf{75.53/0.02} & \textbf{1.80/0.001} & 73.71 &4.14 & 55.08 & 6.95 & 46.27 & 7.02 & 38.96 & 4.25\\

                                & ImageNet~\cite{imagenet} & \textbf{63.16/0.004} & \textbf{11.48/0.05} & 60.32 & 13.04 & 58.57 &18.22 & 45.89 & 15.86 & 49.37 & 3.54\\

    \multirow{2}{*}{VGG16~\cite{vgg}} & MS-COCO~\cite{mscoco} & \textbf{64.64/0.001} & \textbf{4.49/0.003} & 62.88 & 5.00 & 40.03 & 10.07 & 37.94 & 7.06 & 38.96 & 4.25\\
                            & ImageNet~\cite{imagenet} & 58.72/0.03 & 12.23/0.1 & \textbf{59.47} & \textbf{12.66} & 51.59 & 15.75 & 45.75 & 16.23 & 49.20 & 3.13\\
    
    \multirow{2}{*}{InceptionV3~\cite{inception}} & MS-COCO~\cite{mscoco} & \textbf{66.25/0.001} & \textbf{5.88/0.005} & 65.87 & 5.00 & 62.43 & 4.83 & 53.59 & 6.41 & 39.62 & 6.00\\
                            & ImageNet~\cite{imagenet} & \textbf{67.43/0.004} & \textbf{6.34/0.003} & 65.12 & 10.25 & 60.99 & 11.31 & 45.76 & 13.32 & 49.12 & 4.27\\

    \bottomrule
    \end{tabular}
    \label{tab:evaluation-inser-del}
\end{table*}

\subsection{The \textit{pointing game} metrics}
We evaluate the localization capability of saliency map-based approaches for target-specific objects using the pointing game. For the ImageNet dataset, we report the average IOU score for a set of test images across all the models. The IOU score is calculated for the saliency map of the images using the enclosed area of the bounding box, $A_s$ and the annotated segmentation box, $A_t$ for the target-specific input images. Table~\ref{tab:evaluation-pointing} shows the mean IOU score (\%) for the test set of images across all the models and both the datasets. The table also shows the standard deviation of the IOU score for our approach. From the table, it is evident that the performance of our saliency map approach across the models is at least 5\% more accurate than~\cite{rise} and 20\% more accurate than~\cite{lime} at localizing the pixels sensitive towards target-specific classification. Note that we omit the evaluation of the gradient backpropagation technique using the pointing game as the technique highlights only the edges of the target object, which is insufficient for evaluation against this metric.

\begin{table}[ht]
    \centering
    \caption{Mean IOU Score(\%) using pointing game metric}
    \begin{tabular}{lcccccc}
    \toprule
    Model & Ours & RISE~\cite{rise} & GCAM~\cite{grad-cam} & LIME~\cite{lime}\\
    \midrule

    ResNet50~\cite{resnet} & \textbf{79.01/0.02} & 74.9 & 69.11 & 57.29\\
    VGG16~\cite{vgg} &  78.0/0.001 & \textbf{81.12} & 62.31 & 51.17\\
    InceptionV3~\cite{inception} & \textbf{76.0/0.002} & 63.23 & 57.54 & 48.21\\
    \bottomrule
    \end{tabular}
    \label{tab:evaluation-pointing}
\end{table}


\subsection{Convergence}

\begin{figure}[ht!]%
\centering
\subfigure[][]{%
\label{subfig:insertion}%
\includegraphics[width=0.7\linewidth]{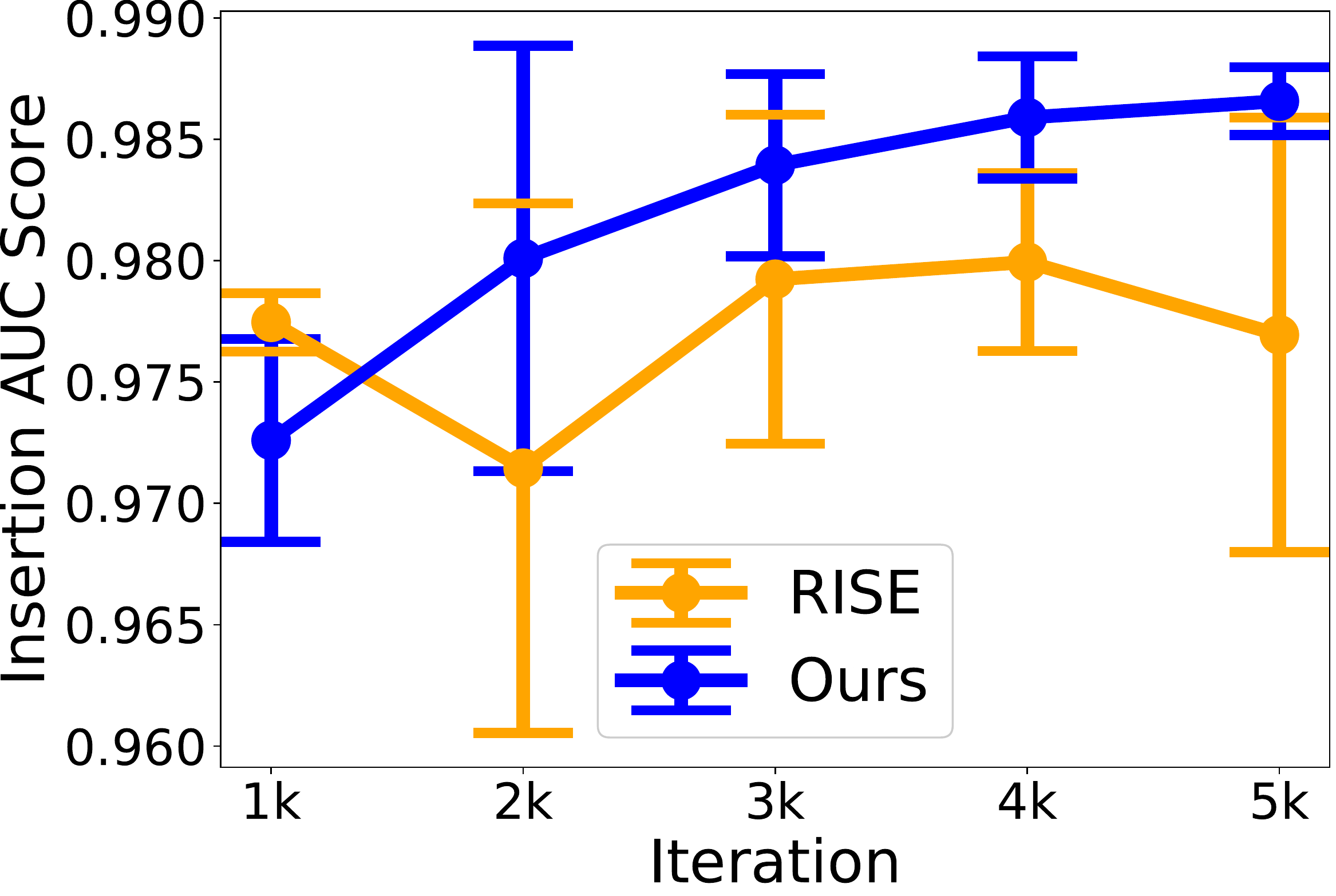}}%
\hspace{8pt}%
\subfigure[][]{%
\label{subfig:deletion}%
\includegraphics[width=0.7\linewidth]{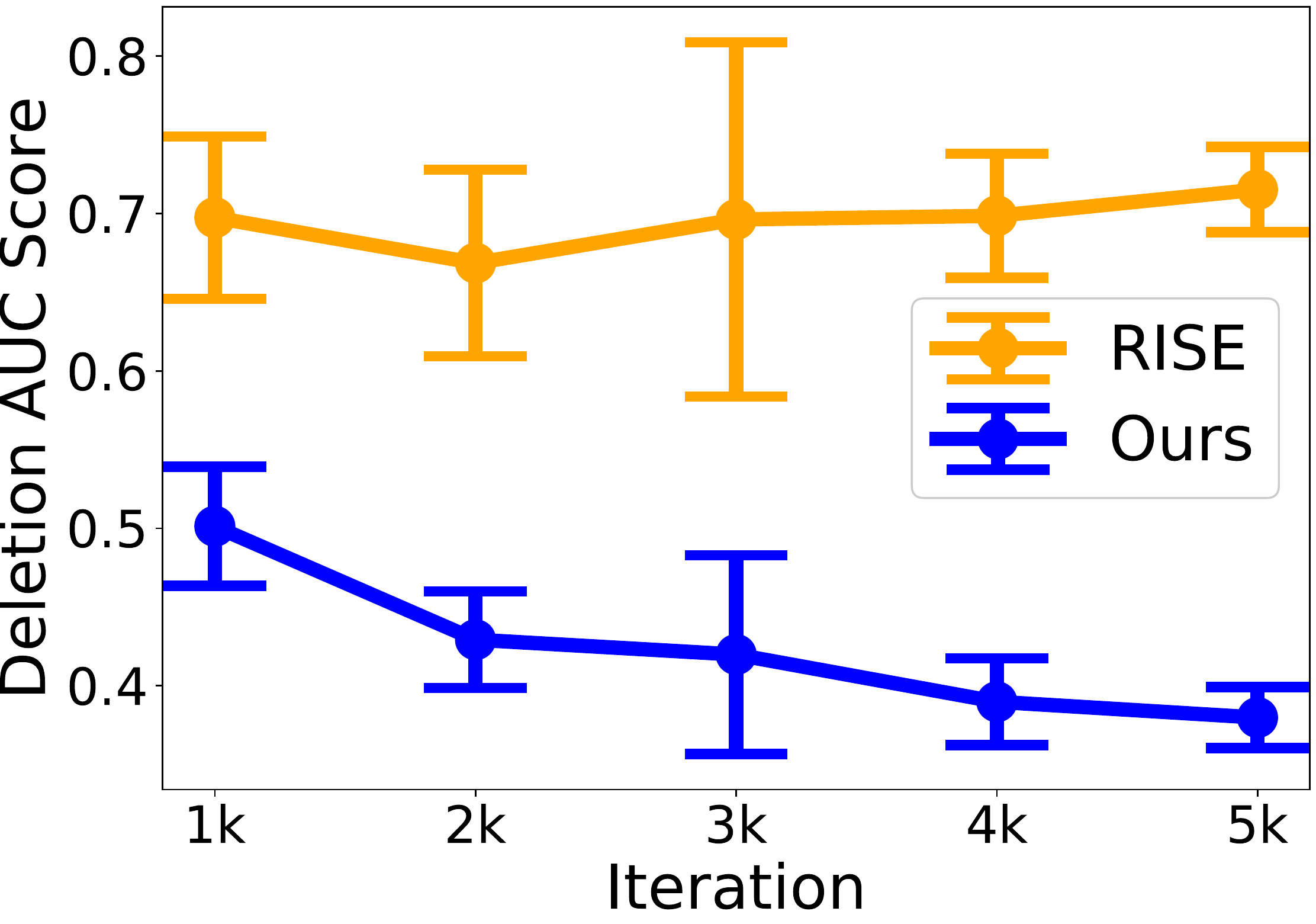}} 
\hspace{8pt}%
\caption{The figure shows the AUC score of insertion~\ref{subfig:insertion} and deletion~\ref{subfig:deletion} for the saliency map of an input image using our approach and RISE~\cite{rise} over the iterations.}%
\label{fig:evaluation-saliency-map}%
\end{figure}

We show that our approach converges to localize pixels sensitive for the target class in less than half the number of iterations as compared to~\cite{rise}. To show this, we calculated the saliency map of a set of randomly selected images from class \textit{honey bee} using our approach and RISE~\cite{rise} for $N=5000$. The criteria for choosing a particular image category includes the presence of at least one distractor object, which makes the target class discrimination difficult. Figure~\ref{subfig:insertion} and~\ref{subfig:deletion} shows the insertion and deletion metric of the saliency map at every iteration. The error bar at each iteration shows the standard deviation of the average AUC. For every $1000^{th}$ iteration, we report the average AUC of insertion and deletion of the pixels using the saliency map generated using both the approaches. From Figure~\ref{subfig:insertion}, we observe that the accuracy increases over iterations and then approaches a point after which updating the saliency mask does not improve the performance of insertion accuracy. This point is the point of convergence. The figure shows that our approach reaches the point of convergence faster than~\cite{rise}. Furthermore, unlike~\cite{rise}, the monotonically increasing and decreasing curve of the mean AUC of insertion and deletion metric shows that the saliency map over iterations generated using our approach is more reliable for decision making. 



\subsection{Evaluation of the variations of the salient region}
In this section, we evaluate the generalizability of our saliency map approach on an input image of a \textit{lynx}. The image is chosen such that the presence of distractor objects makes the discrimination of target class against other classes challenging using a saliency mask. For instance, the image of \textit{lynx} has a background whose pattern and colour match with that of the lynx, which contributes to reducing the target accuracy to 62.15\%. However, we found that the results of the evaluation followed a similar trend for other images evaluated from the test set of the ImageNet dataset. 

\subsubsection{Classification Accuracy of reconstructed images} 
Given a saliency map for an input image of \textit{Lynx}, we report the classification accuracy of a subset of reconstructed images of lynx that are correctly classified as \textit{Lynx} by ResNet50. To obtain the reconstructed images, first, we created a reconstruction mask $R$ by convolving the bounding box for the saliency map of \textit{Lynx} with a kernel of size $(15,15)$. We also trained a GAN~\cite{gan} as the generator (G) using a set of $40$K training images from the ImageNet dataset. The input images were downsampled to a size of $(64 \times 64)$ to speed up the training process of G. The training is followed by feeding G with the input image, the reconstruction mask and a batch of 64 random noise vector $z$ to generate a batch of 64 images. The generated images are used to reconstruct a set of 64 images using Equation~\ref{eq:image-reconstruction}. 

\begin{figure*}[!ht]%
\centering                   
\includegraphics[width=0.8\linewidth]{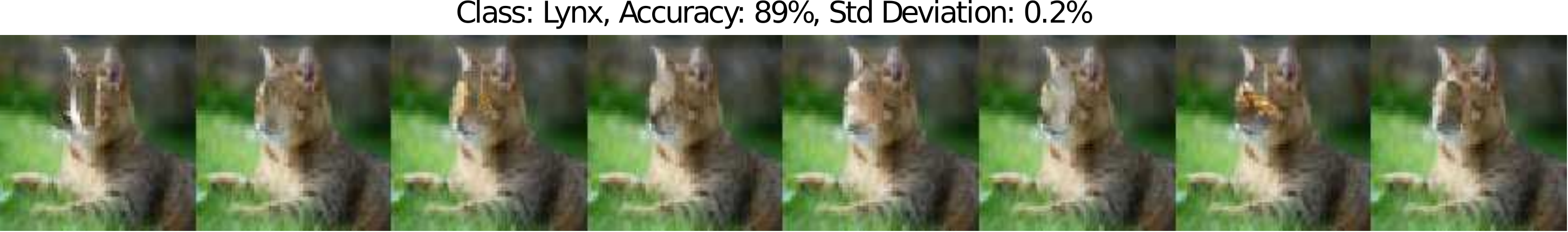}%
\caption{Reconstructed images for the image of a \textit{Lynx} with saliency map ($M$) as shown in Figure~\ref{fig:saliency-mask}}%
\label{fig:reconstruction-results}%
\end{figure*}

Figure~\ref{fig:reconstruction-results} shows a subset of the reconstructed images that are correctly classified to the target class Lynx with an average accuracy of 64\%. The figures are pixelated because the reconstructed images are resized from size $(64 \times 64)$ to size $(224 \times 224)$ where $(64 \times 64)$ is the size of generator output and $(224 \times 224)$ is the size of the saliency map. From the figure, we observe that most of the reconstructed parts of the images (face including eyes, nose, mouth, left side of the face) contain a blob, which is black in the vicinity of mouth and the lower part of the nose. The presence of a consistent feature across the reconstructed images shows that the area around the nose and the mouth of the image is essential for the classification of the image to the class of \textit{Lynx}. 

\begin{figure*}[!ht]%
\centering
\includegraphics[width=0.8\linewidth]{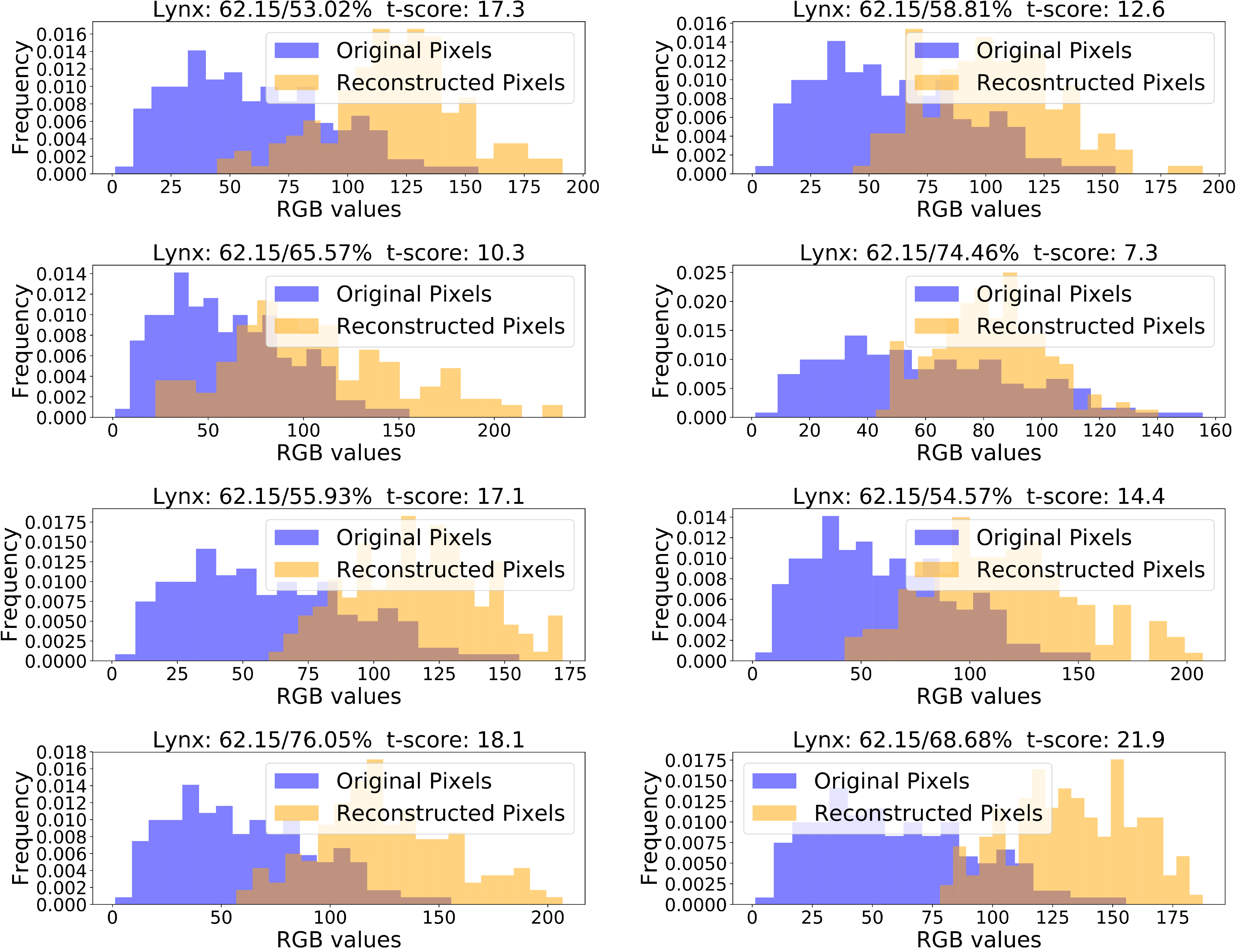}%
\caption{Figures showing the histogram of the reconstructed salient pixels and the original salient pixels. The title of the sub-figures show the accuracy of the class \textit{Lynx} for input image/reconstructed image.}%
\label{fig:reconstruction-histogram}%
\end{figure*}

Figure~\ref{fig:reconstruction-histogram} shows the histograms of the reconstructed salient region of the input and the original salient region of the input. For each of the reconstruction, the figure also shows the accuracy of the target class \textit{Lynx} and the t-score. The t-score is a measure to tell apart the difference between the reconstructed pixels and the original pixels; thus, the higher the t-score value, the larger the differences. Based on the t-score, it is evident that the reconstructed pixels classified to the target class $\textit{Lynx}$ have variations that are significantly different from the original pixels. Thus, these reconstructed pixels of the salient region of the input are the variations, which form alternate explanations for the target class.

\subsubsection{Impact of varying sizes of bounding boxes}
\begin{figure}[!ht]%
\centering
\includegraphics[width=0.7\linewidth]{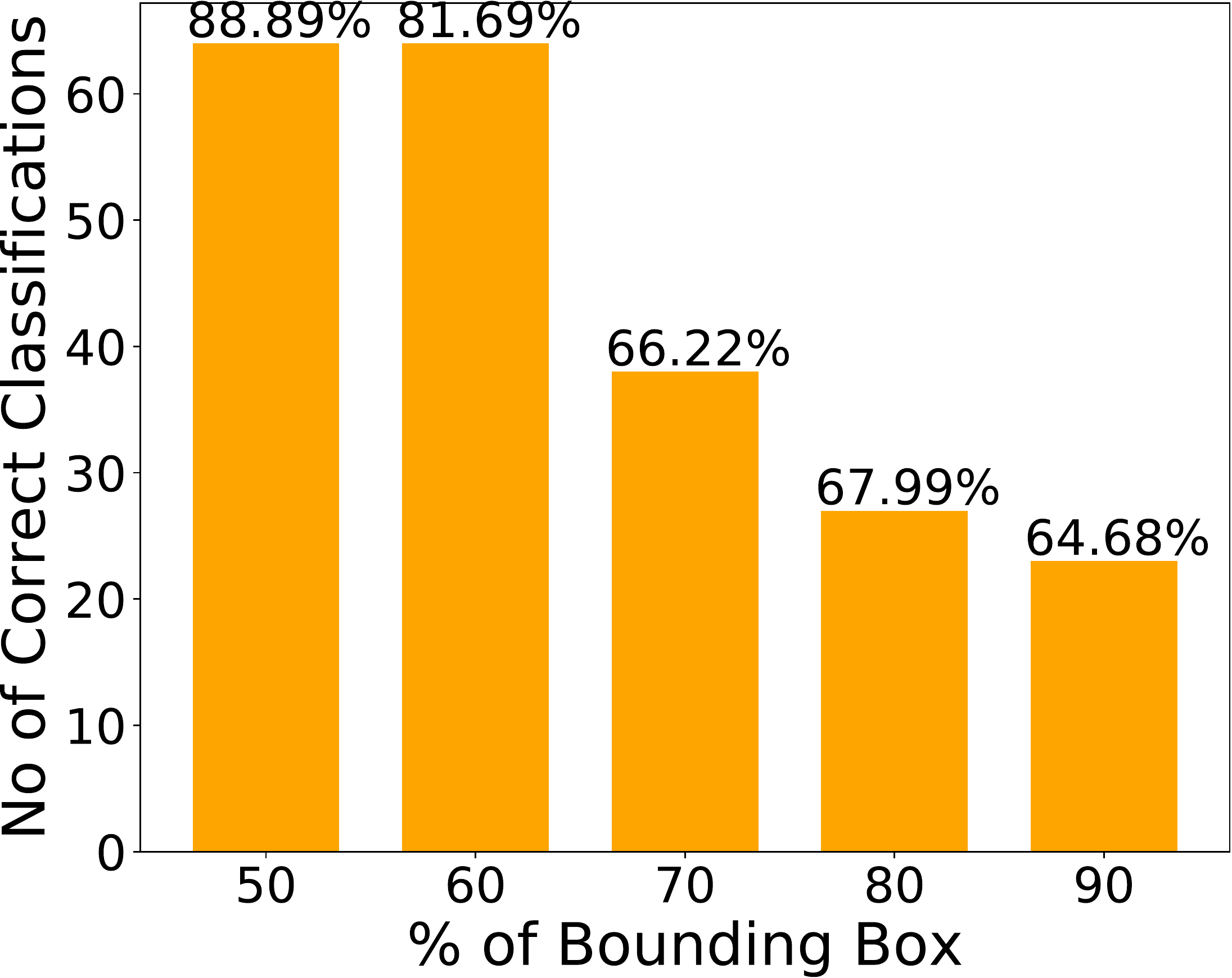}%
\caption{Figure shows the number and accuracy of correct classifications using the reconstructed images over different sizes of bounding boxes.}%
\label{fig:acc-different-size-bounding-boxes}%
\end{figure}

\begin{figure}[!ht]%
\centering
\includegraphics[width=0.7\linewidth]{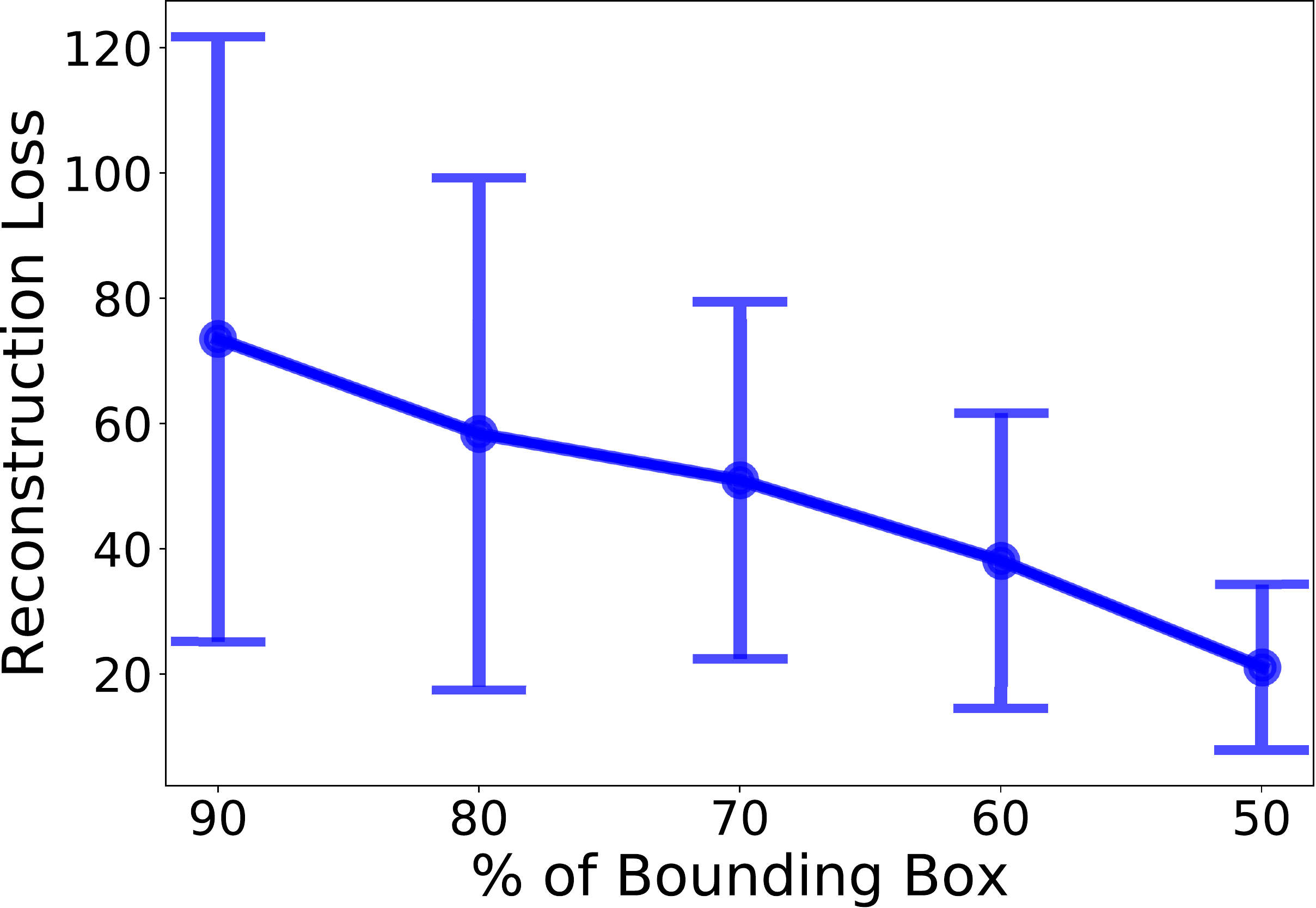}%
\caption{Figure shows the reconstruction loss over the different sizes of bounding boxes.}%
\label{fig:loss-different-size-bounding-boxes}%
\end{figure}

We show that the size of the bounding box enclosing the salient region of the input influences the quality of the reconstructed images and hence the alternate explanations. The idea is that as the size of the salient region to reconstruct shrinks, the evidence from the neighbourhood to reconstruct missing pixels increases, thereby generating contextually similar images. We reconstructed the input image using the varying sizes of the bounding boxes by reducing the bounding box by a factor of $\alpha=0.1$ until half the original size. From Figure~\ref{fig:acc-different-size-bounding-boxes}, it is evident that the number of reconstructed images correctly classified to the target class \textit{Lynx} increases as the size of the bounding box decreases. The most number of correct classifications are observed with the bounding box that is half the size of the original salient region. Equivalently, as shown in Figure~\ref{fig:loss-different-size-bounding-boxes}, the loss incurred by the generator during the reconstruction of the images decreases as the size of the bounding box decrease. This shows that a carefully chosen value of $\alpha$ helps generate contextually similar images. For the image of lynx, a value of $8\alpha$, which retains 80\% of the salient region in the bounding box, generates reconstructed images with an average classification accuracy of 63.4\%.

\section{Conclusion and Future Work}
In this paper, we proposed a generalizable saliency map-based explainability technique for explaining the target-specific outcome of a model. We evaluated our saliency map approach using two datasets: ImageNet and MS-COCO, and against a set of models: VGG16, ResNet50, Inception V3. We also compared our approach against other saliency map-based techniques. Experiments show that our method performs better than the existing explainability approaches across most of the tested scenarios. Furthermore, to evaluate the generalizability of our approach, we evaluated the correctness of the alternate explanations. The accuracy of the reconstructed input shows that the approach can find explanations that are relevant to the classification of input to the target class. 

As part of future work, we will extend our saliency map-based approach for explaining the mode outcome for the input of type time-series. We also plan to use a conditional GAN as the generator. Generator with target class as input will help reconstruct the missing pixels from the salient region using the evidence from a confined neighbourhood of pixels, which will be contextually similar to the salient region of the input, thus, generating realistic-looking salient regions.


\bibliographystyle{IEEEtran}

\bibliography{icpr2020}

\end{document}